\def\year{2018}
\documentclass[letterpaper]{article}
\usepackage{aaai18}
\usepackage{times}
\usepackage{helvet}
\usepackage{courier}
\usepackage{url}  
\usepackage{graphicx}  

\usepackage{amsmath,amsthm,amssymb}
\usepackage{xcolor}
\usepackage{booktabs}
\usepackage{enumerate}
\usepackage[figure,vlined]{algorithm2e}
\newcommand{\Omit}[1]{}
\newcommand{\tup}[1]{{\langle #1\rangle}}
\newcommand{\pair}[1]{{(#1)}}

\newtheorem{definition}{Definition}
\newtheorem{theorem}[definition]{Theorem}

\begin{document}

\title{Planning with Pixels in (Almost) Real Time}
\author{
  Wilmer Bandres \\
  Universitat Pompeu Fabra \\
  Barcelona, Spain \\
  \texttt{wilmer0593@gmail.com}
  \And
  Blai Bonet \\
  Universidad Sim\'on Bol\'{\i}var \\
  Caracas, Venezuela \\
  \texttt{bonet@usb.ve}
  \And
  Hector Geffner \\
  ICREA \& Universitat Pompeu Fabra \\
  Barcelona, Spain \\
  \texttt{hector.geffner@upf.edu}
}

\maketitle

\begin{abstract}
Recently, width-based planning methods have been shown to yield
state-of-the-art results in the Atari 2600 video games. For this, the states
were associated with the (RAM) memory states of the simulator. 
In this work, we consider the same planning  problem but using
the screen instead. By using the same visual inputs, 
the planning  results can  be compared with those of humans and learning methods.
We show   that the planning approach, out of the box and without training,
results in scores that compare well with those obtained by humans and learning methods, and moreover,
by developing an episodic, rollout version of the IW($k$) algorithm, we show that such scores can be obtained
in almost real time. 
\end{abstract}

\section{Introduction}

The Atari 2600 video games supported in the ALE environment \cite{ale}
provide a challenging set of benchmarks for
reinforcement learning (RL) and planning algorithms.
The breakthrough results achieved by the
deep reinforcement  algorithm  DQN  \cite{dqn} made it to the popular press,
and were followed by similar results obtained using  more standard (shallow)
RL methods operating on suitably defined screen  features \cite{shallow}.

The learning and planning settings for ALE deal with the same games but compute  different forms of control.
Learning is done \emph{off-line} and results in  close-loop controllers that require no lookahead.
Planning, on the other hand, is done \emph{on-line:} an action is selected by looking ahead in the future, 
the action is applied, and the process iterates until the game terminates or a maximum number of frames is reached.
The learning approach is slow but the selection of actions,  once the controller has been learned,   is very fast. 
The planning approach, on the other hand works right out of the box without training but results in a selection of
actions that is slower. Another difference is that the   planning approaches
\cite{ale,nir:ijcai2015}, unlike the reported RL approaches, plan with the (RAM) memory states of the Atari
emulator using information that is not available to humans. 

In this work, we address some of the limitations of current planning approaches to the Atari games, so that the resulting
scores can be compared with those achieved
by humans and learning methods. For this,  planning is performed
using the visible information from the screen.
The ``screen states'' are associated with the values of
a  set  of boolean pixel features called B-PROST \cite{shallow}, 
defined and motivated by the neural net architecture  underlying   the successful  DQN algorithm.  
The number of feature bits in B-PROST is more than 
20,000,000, and  the number of possible screen states
is exponential in this number.  Yet, while the complexity of   planning algorithms like breadth-first search
grows  with the number of reachable states, the complexity of
algorithms such as IW(1) \cite{nir:ecai2012} is linear in the number of features.

We show that the IW(1)  planning algorithm using the B-PROST pixel features
turns out to achieve   scores that compare well with those reported for humans
and learning algorithms.
Our main focus, however, will be  on achieving such scores
by planning over very short time windows, approaching real-time play.
For this, we formulate an episodic, rollout version of IW($k$)
and show that it has a much better anytime  behavior than IW($k$). 
Rollout IW($k$) is not a systematic tree-search algorithm like IW($k$),
but  a  sequence of \emph{rollouts} or \emph{episodes} reminiscent of
(open loop) Monte Carlo algorithms, all starting in the same  node, 
each one given by a sequence of actions. The sequence of rollouts
is designed to emulate IW($k$) in polynomial time.
More precisely,  if  a goal has width $k$ \cite{nir:ecai2012},
Rollout IW($k$) will reach the goal in a polynomial number of rollouts.

The empirical results show that Rollout IW(1)
achieves  scores comparable to those of humans and RL algorithms
almost in real-time: decisions are made in 0.5 seconds
using a frameskip of 15  that would require decisions made in  0.25 seconds
for real-time play. The results involve other simple extensions of IW(1) that
are also discussed. 

The paper is organized as follows. We review ALE, IW planning, and the B-PROST
visual feature set, and then consider Rollout IW, extensions, and the experimental
results.

\section{Planning and  Model}

ALE supports a planning and a learning setting. In the planning setting,
the agent performs a lookahead   using the   simulator,
selects and applies  an action, and the process iterates until the game
terminates or a given number of frames is reached.
Often a frameskip of $N$ is used, meaning that decisions are made
every $N$ frames with the  last decision repeated over the following
$N\!-\!1$ frames.

The simulator returns reward (score) information and the goal is to maximize overall reward.  When planning with
memory states, the only planning setting originally supported in ALE, the agent has access to the
system (RAM) state. The lookahead from a state $s_0$ is  aimed  at finding a
sequence of  actions and states  $s_0,a_0,s_1,a_1, \ldots, a_{n-1}, s_n$ with high reward.
The  state at step $i$ is $s_{i+1}=f(a_i,s_i)$ and the reward at step $i+1$ is $r_{i+1}=r(s_{i-1},a_i,s_i)$,
where $f$ is the deterministic state transition function, and $r$ is the deterministic reward function,
both \emph{a priori unknown}. The reward $r_{i+1}$ is observable and
the system states $s_{i+1}$ are observable in the memory setting.
The score obtained along an action sequence is the sum of the observed rewards, and
the discounted score refers to the discounted sum of rewards with
a discount factor slightly less than 1 that favors rewards that are closer in time.
Different planning algorithms manage to explore the value of
different action sequences, and the action selected is the  one that starts
the sequence with highest score. The screen setting is similar except that the
RAM states are not observable. Instead, the screen of the game
is observable but is less informed than the RAM memory  because the former is determined
by the latter and there may be different states associated with the same screen.
The planning task, however, is similar: to find an action sequence that results
in a high score.

\section{Planning with IW}

IW(1) is a simple ``classical''  planning algorithm defined over states that
determine the value  of  a given set of boolean features or atoms $F$ \cite{nir:ecai2012}. Starting with a
given initial state $s_0$, IW(1) is a standard breadth-first search from $s_0$ with one change:
if upon generation of a state $s$, there is no feature $f_i$ in $F$ such that $f_i$ is true in $s$
and false in all the states generated before $s$, the state $s$ is pruned.
\Omit{
CHECK{Implementation is a bit different: just before expansion of $s$, we check whether
  there is a feature $f$ made true by $s$ and not by any previously *expanded* state.
  If so, $s$ is expanded.  Otherwise, $s$ isn't expanded (pruned). Not sure if there
  is a difference but wanted to be clear about it.}
  }
In other words, the only states that are not pruned
are those that make \emph{some} feature in $F$ true for the first time.
Such states are said to have novelty $1$. 
The number of states expanded  in the IW(1) search
is thus  \emph{linear} in the size of the feature set
and not in the number of states as in  standard breadth-first search.
The algorithm IW($k$) is the algorithm IW(1) but applied to
the larger feature set $F^k$ made up of the  conjunctions
of $k$ features from $F$.

In classical planning, the set $F$ of features commonly used is the
set of atoms for the task. Basic properties of IW($k$) 
are that many planning benchmark domains have a bounded and
small \emph{width} $w$ no greater than 2 when goals are single atoms, and
that such instances are solved optimally (shortest plans) by running IW($w$)
in low polynomial time in the number of atoms.
For example, the goal $on(x,y)$ for any two blocks $x$ and $y$ in Blocksworld
can be shown to have width no greater than $2$, no matter the number of blocks
or initial configuration. This means that IW(2) finds an optimal (shortest)
plan to achieve $on(x,y)$ in polynomial time in the number of atoms and blocks
even though the state space for the problem is exponential in both. 
The same is true for most classical planning  benchmarks.
The majority of the benchmarks, however, do not feature a single atomic goal
but a conjunction of them, and a number of extensions of IW($k$) have been
developed aimed at problems with conjunctive goals, some of which represent
the current state of the art in classical planning \cite{nir:aaai2017,guillem:ijcai2017}.
Since these algorithms can work with simulators, variations of IW($k$) have
been used in other settings such as the general-video game competition
\cite{tomas:aiide2015}, and  task and motion planning problems in robotics
\cite{jonathan:2017}.

\section{Planning  with Memory  States}

Among the  original planning baselines,
random playing and breadth-first search
achieve very  low scores, and only UCT \cite{uct}
does well.  IW(1) was shown to improve the performance
of UCT using the 128 bytes of memory as features \cite{nir:ijcai2015}.
The alternative representation where the boolean features
are associated with each one of the  1024 bits in memory
makes  IW(1)  runs  faster but with poorer results.
The IW algorithms  are purely exploratory  and do not take rewards into account
for focusing the search. In order  to improve the scores achieved by IW(1),
two variants  were proposed,  2BFS, a best-first search algorithm
\cite{nir:ijcai2015}, and prioritized IW(1)  \cite{carmel:ijcai2016},
that managed to improve the scores substantially.  From a theoretical point of view,
however, unlike  IW($k$), neither 2BFS nor prioritized IW(1) run in polynomial time in the number of features. 

\section{Planning with Pixels}

The sensory input in ALE is given by images defined by a pixel array 160 wide and 210 high, with 
pixels that  may have up to 128 colors. In principle, we can use the resulting set of
$160 \times 210 \times 128$ booleans
as the set of features in IW, yet
the performance of IW, like the performance of learning algorithms,
is sensible to the choice of features. Features that capture
meaningful structure  normally yield better results than raw
features that do not. For this reason, screen pixels are mapped  into the
set of visual features B-PROST defined by \citeauthor{shallow} for ALE, 
which are motivated by the design choices in the DQN architecture. 

The Atari screen is split into $16 \times 14$ disjoint tiles, each comprised of  $10 \times 15$  pixel patch.
For each tile $(i,j$) and color $c$, there is a \emph{basic feature}
$f_{i,j,c}$ which is 1 (true) iff the image contains a
pixel of color $c$ in the tile $(i,j)$.
The number of basic features is $\text{28,672} = 16 \times 14 \times 128$.

The basic pairwise relative offsets in space  features (B-PROS)
track relative distances among pairs of basic features.
A  B-PROS feature $f_{i,j,c,c'}$ is 1  iff there is a  pixel of color $c$ in tile $t$ and a pixel of color $c'$ in tile
$t'$ such that the horizontal and vertical offsets of  $t'$ in relation to $t$
are $i$ and $j$, for $-13 \leq i \leq 13$ and $-15 \leq j \leq 15$.
The number of non-redundant B-PROS features is
$\text{6,856,768} = (\text{28,672} + ((31 \times 27 \times 128^2 - 128)/2 + 128))$.

Finally, the basic pairwise relative  offsets in time    features (B-PROT)
represent pairwise relative offsets between basic  features obtained from the screen 
at two  different time points: the current and previous decision times (recall that decisions are made at  intervals given by the frameskip parameter).
More specifically,  the B-PROT feature $f'_{i,j,c,c'}$ is 1 iff  there is a  pixel of color $c$ in  tile $t$
at the previous decision point and  a pixel of color $c'$ in tile $t'$ at the current decision point
such that the offsets of $t'$ in relation to $t$  are $i$ and $j$. B-PROT features are non-Markovian,
and hence the ``screen state'' contains information about the previous screen too.
The number of B-PROT features is $\text{13,713,408} = 31\times27\times128^2$, which is twice the number of B-PROS features as there are no redundant pairs. 

The \textbf{B-PROST} feature set contains the Basic, B-PROS, and B-PROT sets for a total of
20,598,848 features. \citeauthor{shallow} go on to introduce additional features aimed at grouping pixels
into object-like blobs. While in our results we compare with their RL algorithm working on their
full set of  Blob-PROST features, we stick to the  simpler B-PROST set.
An  analysis of the impact of different visual features  on the performance of the
planning algorithms would be interesting but is beyond the scope of this work.

Two final comments about the use of the visual features. Following
\citeauthor{shallow} and previous work,  background pixels
are removed from the screen, a transformation that is like
removing ``static'' atoms from a planning problem.
A background pixel is one that preserves its color in all
reachable screens and it is ``removed'' by ``painting''
it with a common ``background''  color. This reduces the number
of active features (features with value $1$).
Background pixels are identified dynamically as the game
evolves. A first set of background pixels is identified by
performing 100 random actions. Then, every time a screen is
scanned for computing the features of a state, the background
status of each pixel is updated by comparing the pixel value
with a stored value.

The ``states'' in IW(1) and the variants introduced below refer
to the \emph{state of the B-PROST features}.
There are potentially $128^{160\times210}$ different screen
configurations but IW(1) is guaranteed to expand 20,598,848
states at most, but usually much fewer as the features are
sparse and most combinations of them never become true.


\section{IW Rollouts}

While IW(1) is a low polynomial time algorithm whose effectiveness has been shown in classical planning and games, 
its suitability for real-time game playing, where actions have to be taken in a few hundred milliseconds, 
is limited by  the underlying breadth-first search.  IW(1) can  search much deeper than
breadth-first search over a limited time window,  but if the window is small,
nodes that are beyond a certain depth  will not be explored either.
Since a main  goal of this work is to play the Atari 2600 games in real time,
we introduce an alternative to IW(1), called \textbf{Rollout IW(1)}, that does not have this limitation and has a
much better anytime behavior.
In addition, Rollout IW(1) asks less from the simulator: while tree search algorithms like IW(1) need the facility of
expanding nodes, i.e., of applying all actions to a node, the rollouts in Rollout IW(1) apply
just one action per node.  We will prove  that Rollout IW(1) terminates
in polynomial time, reaching every  width 1 goal  optimally (length-wise) like IW(1).
While the description and the results below are for Rollout IW($k$) with $k=1$, 
the generalization to higher values of $k$ is direct: for this, the set of boolean features $F$ is 
replaced by the larger set $F^k$ of features comprised of  the  conjunctions  of $k$ features from $F$.

Given a root node defined by an  initial state, a sequence of rollouts starting at the root node
determines a tree. A rollout is an action sequence, and a node in the tree stands
for the  set of state-action trajectories generated by the rollouts that
share  a common prefix and end with the same  state, which is the state
associated with the node. The edges of the tree are the actions. 
For the sequence of rollouts to  replace the breadth-first construction of the tree,
Rollout IW(1) considers ``extended features''  $\pair{f,d}$ where $f$ is a feature in $F$ and $d$ is an integer in
$0 \leq d \leq |F|$. A trajectory  $\tau$ ending in a state $s$ makes the extended feature (e-feature) $\pair{f,d}$ true
iff $s$ makes $f$ true, and the length of the path as measured by the number of actions in $\tau$ is $d$.
In addition, the min $d$ value over all the e-features $\pair{f,d}$ reached is stored as $d[f]$
for each feature $f \in F$. This  value  is initially $0$ for features $f$ true in the root state
and $\infty$ otherwise. 

\begin{algorithm}[t]
  \SetKw{Break}{break}
  \SetKw{Continue}{continue}
  \DontPrintSemicolon
  Generate-Lookahead-Tree: \\
  \Begin{
    $d := \text{make-empty-table}()$ \\
    $r := \text{make-root-node}()$ \\
    \While{$r$ isn't solved and still within budget}{
      $\text{Rollout}(r, d)$
    }
  }

  \BlankLine

  $\text{Rollout}(n, d)$: \\
  \Begin{
    $\text{Fill-Node}(n)$ \\
    \While{$n$ isn't solved}{
      \lIf{$n$ isn't expanded}{
        $\text{Expand}(n)$
      }
      $n := \text{Pick random unsolved child of $n$}$ \\
      $\text{Fill-Node}(n)$

      \BlankLine
      \uIf{$n$ is terminal}{
        $n.visited := \textbf{true}$ \\
        $\text{Solve-and-Propagate-Label}(n)$ \\
        \Break
      }
      \Else{
        $f := \text{Get-Novel-Feature}(n, d)$ \\
        \uIf{$n.depth < d[f]$}{ 
          \emph{/* case 1 */} \\
          $n.visited := true$ \\
          $\text{Update-Feature-Table}(d, n)$
        }
        \uElseIf{$\neg n.visited \,\land\,n.depth \geq d[f]$}{ 
          \emph{/* case 2 */} \\
          $n.visited := true$ \\
          $\text{Solve-and-Propagate-Label}(n)$ \\
          \Break
        }
        \uElseIf{$n.visited \,\land\,d[f] < n.depth$}{ 
          \emph{/* case 3 */} \\
          $\text{Solve-and-Propagate-Label}(n)$ \\
          \Break
        }
        \Else{
          \emph{/* case 4: continue extending rollout */} \\
        }
      }
    }
  }
  \caption{Pseudo-code of Rollout IW(1).}
  \label{fig:code}
\end{algorithm}

\Omit{
\begin{figure}[t]
\begin{center}
\begin{verbatim}
Initially (before first rollout only)
 Initial novelty table d[f]=infty for all f
 Tree contains root node only 
ROLLOUTS: 
  current action prefix AP = empty ;
   s= s_0
- Loop:
  For current action prefix AP 
  Extend AP with any  action ai+1 s.t.
  AP' = AP,ai+1  *not labeled SOLVED*,
  and let s be resulting  state
  /* Four cases * /
1 If there is  (f,d) in s with d < d[f]
   THEN
    update novelty table  with new atoms 
     Set AP := AP'
     Add AP' to Tree of action sequences ..
     Go back to LOOP
2. ELSE if AP' is *not* in Tree
     and *no new atom* in s,
   THEN  mark AP' SOLVED and PRUNE
      Backpropagate SOLVED label UP
      exit LOOP
3  ELSE if AP'*in* Tree and no f  d=d[f],
     THEN  mark AP' SOLVED and PRUNED
     Backpropagate SOLVED label UP
      exit LOOP
4. ELSE (if AP'  in Tree, d=d[f] for some f ..
    Set AP := AP'
    Add AP' to Tree; AP' cannot have beein
    Go back to LOOP
END LOOP
\end{verbatim}
\end{center}
\caption{Pseudo Code of Rollout IW(1). **Revise: Better explanation in text. This is old**}
\end{figure}
}

Nodes in the tree can be labeled  as SOLVED. Initially, the tree contains  the root node only and the 
algorithm terminates when this node is SOLVED or the time budget for selecting an action is over.
In order to define the  Rollout IW(1) algorithm fully (pseudo-code shown in Figure~\ref{fig:code}),
four aspects must be made precise: A)~how actions are selected in rollouts,
B)~how rollouts are terminated, C)~how nodes  are labeled SOLVED, and D)~how labels  propagate  up in the tree. 

Rollouts start in the root node, and recursively extend a path prefix $\tau'$  with an action $a$ and a node $n$ representing a state $s$.
The resulting path $\tau = \tup{\tau',a,n}$  may already be in the tree or not: if it is in the tree,  the only condition in the choice of the
action $a$ is that node $n$ is \emph{not} labeled SOLVED. A rollout can   select \emph{any action}  in  a node 
as long as it  does not lead to a SOLVED node.  Four cases are distinguished after  extending a rollout with a new action.
In two of these cases the rollout is terminated, the tip nodes are labeled  SOLVED, and the label is propagated up the tree.
In the other two cases, the rollout is extended with a new action and the process iterates.
If $\tau=\tup{\tau',a,n}$ is the tree branch that results from extending a rollout with an
action $a$, the four cases are: 

\begin{enumerate}[$\bullet$]
  \item Case 1: if node $n$ is new in the tree and makes some e-feature $(f,d)$ true with $d < d[f]$, \textbf{continue} rollout.
  \item Case 2: if node $n$ is new in the tree but $d \geq d[f]$ for each e-feature $(d,f)$ true in $n$, \textbf{terminate} rollout.
  \item Case 3: if node $n$ is already in the tree but for no e-feature $(f,d)$ true in $n$, $d = d[f]$, \textbf{terminate} rollout.
  \item Case 4: if node $n$ is already in the tree and makes some e-feature $(f,d)$ true with $d = d[f]$, \textbf{continue} rollout, 
\end{enumerate}

Cases 1 and 2 involve a  new branch in the tree, leading to a new node. The rollout terminates in the node, if the node  does not bring
a ``new'' extended feature, which is a pair $(f,d)$ that improves the distance to $f$, i.e., with  $d < d[f]$.
Cases 3 and 4 are different. Since Rollout IW(1) is based on rollouts and not on  tree search like IW(1), it means
that it has to search multiple times beneath a node $n$ in order to explore each of its children.
In such cases, $n$ is not a new node in the tree but an existing  node.
The  search beneath such  nodes is  terminated when for all e-features $(f,d)$ true in $n$,
the distance $d$ to $f$ can be shown not to be the shortest  given the search so far.

In Case 1, the table $d$ that tracks the shortest distances to features $f$ is updated
by setting $d[f]$ to  the minimum between  $d[f]$ and the depth of  $n$, for each feature $f$ true in $n$. 
In Cases 2 and 3, the last  node $n$ of the rollout is labeled as SOLVED and the label is propagated
up the tree. Nodes  in the tree become  SOLVED when all of their children have been
generated and all of them have been labeled as SOLVED.

In the code shown in  Figure~\ref{fig:code}, the procedure 
Get-Novel-Feature() returns \emph{any}  feature $f$ for a pair $(f,d)$ true in $n$
with $d < d[f]$,  if one exists, else  \emph{any}  feature $f$ with  $d=d[f]$ if one exists,
and any feature otherwise.  The Fill-Node() procedure fills up the data structure and
calls  the simulator except when the state is cached (see below). 
A basic property of the sequence of rollouts is:

\begin{theorem}
\label{thm:1}
Each rollout improves the distance $d$ to some feature $f$, i.e.\ it reaches an e-feature $(f,d)$
with $d < d[f]$, or labels the child of a node in the tree as SOLVED.
\end{theorem}


The length of rollouts is bounded by $|F|$ as for every non-tip node, 
$d \leq d[f]$ must be true for some feature $f$, yet $d$ grows monotonically
along the rollout while $d[f]$ doesn't  grow. Similarly, the number of nodes in the tree is always  bounded by the total number
$|F|^2$ of e-features $(f,d)$,  as each new node must improve $d[f]$ for some feature $f$.
Theorem~\ref{thm:1} implies then that:

\begin{theorem}
Rollout IW(1) terminates (root node SOLVED) in a number of rollouts bounded by
$|F|^2 \times b$ where $b$ is the branching factor (maximum number of
actions that can be applied at a give node).
\end{theorem}

Since IW(1) can generate up to $|F| \times b$ nodes, Rollout IW(1) involves an  overhead. 
Rollout IW(1) does not compete with plain IW(1) when both are run to completion but
it is an appealing alternative over short time windows. 

In order to characterize the states that are reached by Rollout IW(1) from a given state $s_0$,
we appeal to the notion of \emph{width} from \cite{nir:ecai2012}:
a goal $G$ defined from the features in $F$ has width 1 iff
there is a trajectory $s_0,a_0,\ldots,a_{n-1},s_n$ where $G$ is true in $s_n$,   $n > 0$,
such that for each state $s_i$ in the trajectory:
1)~the prefix $s_0,a_0,\ldots,s_i$ is optimal   for some feature $f_i$ (no other trajectory reaches $f_i$ with less number of actions), 
2)~any  optimal trajectory to $f_i$ can be extended into an optimal trajectory for  $f_{i+1}$, 
and 3)~the optimal plans for $f_n$ are optimal for $G$. IW(1) is guaranteed to reach all width 1 goals no matter how the children nodes are ordered
in the underlying breadth-first search. The same is true for Rollout IW(1):

\begin{theorem}
Rollout IW(1) is guaranteed to reach every width 1 goal in time that is polynomial in the number of features.
\end{theorem}

The practical value of Rollout IW(1), however,  is its \emph{anytime behavior} that
follows from replacing the breadth-first search in plain IW(1) by depth-progressing
rollouts. As mentioned above, all definitions and properties carry easily
to Rollout IW($k$) for $k > 1$. 


\section{Extensions and Variations}

Before testing IW(1) and its rollout version, we consider one optimization and two variations.

\subsection{Partial Caching}

On-line planning schemes for the Atari games have appealed to different ways of using  previous lookahead results
in the current lookahead \cite{ale}. Indeed, once an action has been selected and applied, the subtree rooted
at the selected child contains information that can be used ``for free''  in the new lookahead. 
In the IW(1) planning scheme of \citeauthor{nir:ijcai2015} (\citeyear{nir:ijcai2015}), that operates over
the RAM states, if the lookahead tree in a given situation involves the trajectory $s_0,a_0,s_1,a_1,\ldots,s_n$
and the action $a_0$ is selected and applied, the sub-branch $s_1,a_1, \ldots,s_n$ is used for both
avoiding simulator calls (it can be predicted that the action sequence $a_1,\ldots, a_{i-1}$ will lead to
the state $s_i$), and for adding extra states to the search (states along such cached branches are not
pruned and are not used for updating the novelty table).  In the screen setting, we do the same but replace
the RAM states with screen states. There is however an important change: if $s_1,a_1,\ldots,s_n$ is a cached
path made of screen states rather than memory states, then predicting the next screen state $s_{i+1}$
after an action $a_n$ when the path $s_1,a_1,\ldots,s_n,a_n,s_{n+1}$ is not cached may involve
up to $n$ simulator calls because the screen state $s_{n+1}$ can only be obtained by using
the simulator from the root state.
In order to avoid these repeated  calls we assume an ``intelligent'' simulator that caches such states within
the same lookahead tree, while keeping them hidden from the planning agent.

\subsection{Penalties and Risk Aversion}

In the ALE setting, the planning agent observes when he dies, and there are games, such as
Breakout, on which the loss of a life does not generate a negative reward (penalty).
In other games, positive and negative rewards are not calibrated. For example, in
Space Invaders, losing a life  results in a reward of -1 while shooting
an  ``alien'' results in  a reward  of 5 or higher. This results in  situations
where the lookahead does not find sufficiently high rewards, or the planning agent may select
actions leading to an early and unnecessary death. To avoid this,  negative rewards
are multiplied by a large constant $\alpha=\text{50,000}$ and a high negative reward  of $-10\alpha$ is
used for deaths.  Since this form of ``reward shaping'' appears to be fair for comparing with humans
but less so for comparing with learners, we report results with and without this ``risk averse'' transformation.

\subsection{From Subgoaling to Subscoring}

The IW($k$) algorithm is effective for achieving atomic goals in classical planning but
not for achieving sets $G$ of atomic goals \cite{nir:ecai2012}. A variation of IW($k$),
that we call $\textrm{IW}_G(k)$, provides a simple mechanism for doing that, and will
serve as the basis for making a score-sensitive IW algorithm that is also polynomial.
In the implementation of IW(1), a state $s$ has novelty 1 iff there is a true feature $f$
in $s$ that is not marked as ``reached''  in a \emph{novelty table} that tracks the
features that have been reached so far.
$\textrm{IW}_G(1)$ is like IW(1) but with one change: there are $|G|$ novelty tables,
with the $k$-th table tracking the features that have been reached by the states that
make exactly $k$ goals true \cite{nir:aaai2017}.
A state $s$ has novelty 1 in $\textrm{IW}_G(1)$ iff there is a true feature $f$ in $s$
that is not marked as ``reached'' in the $k$-th novelty table where  $k=\#g(s)$ is the
number of goals that are true in $s$.
Like IW(1), $\textrm{IW}_G(1)$ is a breadth-first search but while IW(1) expands up to
$|F|$ nodes,  $\textrm{IW}_G(1)$ can expand up to $|F| \times |G|$ nodes. 

IW(1) with subscoring, denoted as $\textrm{IW}_S$(1) is similar to IW(1) with subgoaling,
namely $\textrm{IW}_G$(1), but with the goal counter $\#g(s)$ replaced by the integer-valued
$logscore(s)$ whose value approximates the logarithm base 2 of the score (reward) accrued
on the \emph{unique path} leading to state $s$.\footnote{Rollout algorithms explore a
  lookahead tree and thus each state (node) in the tree corresponds to a path
  in the lookahead tree.
}
A state has novelty 1 in $\textrm{IW}_S$(1) iff there is a feature $f$ true in $s$ that
is not marked as reached in the $k$-th table for $k=\text{logscore}(s)$.

The scoring function is as follows. If the reward $r(s)$ of the path leading 
to $s$ is non-positive, $\text{logscore}(s)=0$.
Else, if $r(s)<1$, $\text{logscore}(s)=\lfloor\log_2(r(s))\rfloor$.
Else, if $r(s)\geq 1$, $\text{logscore}(s)=1 + \lfloor\log_2(r(s))\rfloor$.
Hence, all non-positive path rewards are associated with the index 0, all
rewards in $(0,1)$ are associated with negative integer indices, and all
path rewards in $[1,\infty)$ are associated with positive integer indices.
It is not difficult to show that under the assumption of bounded reward per
step, only a polynomial number of novelty tables are needed when doing rollouts
of either polynomial or exponential depth, and this is independent of the number
of such rollouts.

\section{Experimental Results}

The benchmark contains 58 games for the Atari 2600, all with screens of $160\times210$
pixels.
Experiments were performed on an Amazon EC2 cluster made of m4.16xlarge
instances each featuring 64 Intel Xeon E5-2686 CPUs running at 2.30GHz
and 256Gb of RAM.
IW(1) and three versions of Rollout IW(1) running over the B-PROST features
are compared with the shallow RL algorithm using the full Blob-PROST feature
set \cite{shallow}, the DQN algorithm, the results of the human player
reported in \cite{dqn}, and the IW(1) algorithm over the RAM states \cite{nir:ijcai2015}.
The last one only included as a reference.
The algorithms are evaluated with time budgets for online decision making of
0.50 and 32 seconds, and a frameskip of 15 that is compatible with human play.
In addition, for actions that do not manage to change the value of any B-PROST
feature in 15 frames, we apply the action for another 15 frames before pruning
the node. Each data point in the results is the average of 5 executions.
We denote with RA Rollout IW(1) and RAS Rollout IW(1) the risk-averse and
risk-averse plus subscoring versions of Rollout IW(1), respectively.

Table~\ref{table:narrow} shows how RAS Rollout IW(1) with a time window of half
a second performs as good as the human in 25 games (51.0\%), and achieves 75\%
of the human scores in 29 games (59.1\%), sometimes overperforming the human by
a wide margin.
RAS Rollout IW(1) in half a second is also better than the human, DQN, and the
shallow RL algorithm in 15 games (30.6\%).
The human player, DQN, and the RL algorithm are best in 16, 12, and 6 games
respectively (32.6\%, 24.4\%, and 12.2\%).

\begin{figure}[t]
  \centering
  \includegraphics[width=.95\columnwidth]{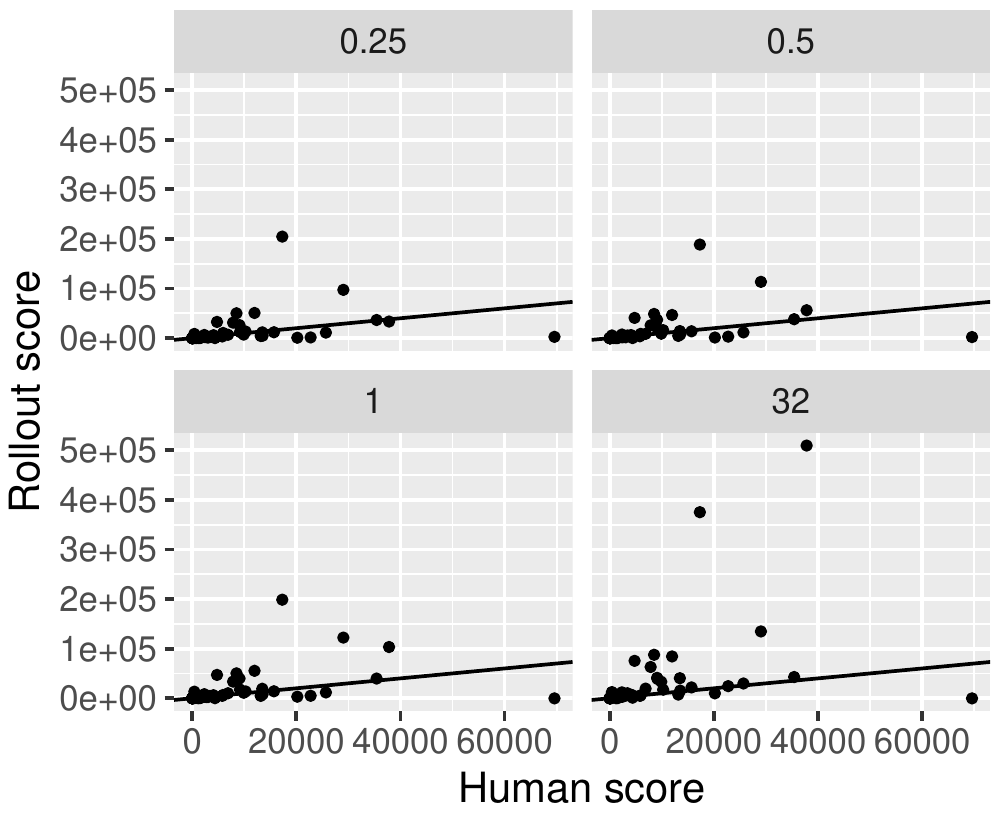}
  \caption{Scatter plots for RAS Rollout IW(1) with time budgets of 0.25, 0.50, 1, and 32
    seconds vs.\ human playing over 49 games.
    Points on or above the diagonal correspond to games on which RAS Rollout IW(1) is as
    good as the human player.
  }
  \label{fig:110:scatter}
\end{figure}

The scatter plots in Figure~\ref{fig:110:scatter} depict the performance of RAS
Rollout IW(1) with different time windows in relation to human playing.
Each point on or above the diagonal stands for a game on which RAS Rollout-IW(1)
is at least as good as the human player. A truly real-time configuration corresponds
to a time budget of 0.25 seconds, as the agent plays 4 times per second when using
a frameskip of 15.

\begin{table}[t]
  \centering
  \resizebox{\columnwidth}{!}{
\newcommand{\B}{\color{red}}
\begin{tabular}{@{}rrrrr@{}}
\toprule
                               &              &               &                 &    RAS Rollout IW(1) \\
                          Game &        Human &           DQN &   RL-Blob-PROST &          budget 0.5s \\
\midrule                                                               
\large                   alien &      6,875.0 &       3,069.0 &         4,154.8 &\bf\B         8,550.0 \\ 
\large                  amidar &\bf   1,676.0 &         739.5 &           408.4 &              1,161.0 \\ 
\large                 assault &      1,496.0 &\bf\B  3,359.0 &         1,107.9 &                264.6 \\ 
\large                 asterix &      8,503.0 &       6,012.0 &         3,996.6 &\bf\B        48,700.0 \\ 
\large               asteroids &\bf  13,157.0 &       1,629.0 &         1,759.5 &              4,486.0 \\ 
\large                atlantis &     29,028.0 &\B    85,641.0 &\B      37,428.5 &\bf\B       113,460.0 \\ 
\large              bank heist &\bf     734.4 &         429.7 &           463.4 &                268.0 \\ 
\large             battle zone &     37,800.0 &      26,300.0 &        26,222.8 &\bf\B        56,200.0 \\ 
\large              beam rider &      5,775.0 &\bf\B  6,846.0 &         2,367.3 &              3,729.2 \\ 
\large                 bowling &\bf     154.8 &          42.4 &            65.9 &                 51.6 \\ 
\large                  boxing &          4.3 &\B        71.8 &\bf\B       89.4 &\B               78.6 \\ 
\large                breakout &         31.8 &\bf\B    401.2 &\B          52.9 &\B               79.8 \\ 
\large               centipede &     11,963.0 &       8,309.0 &         3,903.3 &\bf\B        46,661.4 \\ 
\large         chopper command &\bf   9,882.0 &       6,687.0 &         3,006.6 &              8,900.0 \\ 
\large           crazy climber &     35,411.0 &\bf\B114,103.0 &\B      73,241.5 &\B           38,120.0 \\ 
\large            demon attack &      3,401.0 &\bf\B  9,711.0 &         1,441.8 &\B            5,201.0 \\ 
\large             double dunk &        -15.5 &         -18.1 &\B          -6.4 &\bf\B            -4.0 \\ 
\large                  enduro &\bf     309.6 &         301.8 &           296.7 &                137.4 \\ 
\large           fishing derby &\bf       5.5 &          -0.8 &           -58.8 &                -61.8 \\ 
\large                 freeway &         29.6 &\B        30.3 &\bf\B       32.3 &                  3.6 \\ 
\large               frostbite &\bf   4,335.0 &         328.3 &         3,389.7 &              1,494.0 \\ 
\large                  gopher &      2,321.0 &\bf\B  8,520.0 &\B       6,823.4 &\B            7,256.0 \\ 
\large                gravitar &\bf   2,672.0 &         306.7 &         1,231.8 &              2,410.0 \\ 
\large                    hero &\bf  25,673.0 &      19,950.0 &        13,690.3 &             11,480.0 \\ 
\large              ice hockey &          0.9 &          -1.6 &\bf\B       14.5 &\B                5.2 \\ 
\large              james bond &        406.7 &\B       576.7 &\B         636.3 &\bf\B         5,340.0 \\ 
\large                kangaroo &      3,035.0 &\bf\B  6,740.0 &\B       3,800.3 &              1,800.0 \\ 
\large                   krull &      2,395.0 &\B     3,805.0 &\bf\B    8,333.9 &              1,645.2 \\ 
\large          kung fu master &     22,736.0 &\B    23,270.0 &\bf\B   33,868.5 &              2,980.0 \\ 
\large     montezuma's revenge &\bf   4,367.0 &           0.0 &           778.1 &                100.0 \\ 
\large               ms pacman &\bf  15,693.0 &       2,311.0 &         4,625.6 &             13,746.8 \\ 
\large          name this game &      4,076.0 &\bf\B  7,257.0 &\B       6,580.1 &\B            6,128.0 \\ 
\large                    pong &          9.3 &\B        18.9 &\bf\B       20.2 &                 -1.4 \\ 
\large             private eye &\bf  69,571.0 &       1,788.0 &            33.0 &              2,160.0 \\ 
\large                   qbert &     13,455.0 &      10,596.0 &         8,072.4 &\bf\B        14,160.0 \\ 
\large               riverraid &\bf  13,513.0 &       8,316.0 &        10,629.1 &              7,138.0 \\ 
\large             road runner &      7,845.0 &\B    18,257.0 &\B      24,558.3 &\bf\B        25,780.0 \\ 
\large                robotank &         11.9 &\bf\B     51.6 &\B          28.3 &\B               30.0 \\ 
\large                seaquest &\bf  20,182.0 &       5,286.0 &         1,664.2 &              1,236.0 \\ 
\large          space invaders &      1,652.0 &\bf\B  1,976.0 &           844.8 &\B            1,812.0 \\ 
\large             star gunner &     10,250.0 &\bf\B 57,997.0 &         1,227.7 &\B           15,960.0 \\ 
\large                  tennis &         -8.9 &\B        -2.5 &\B           0.0 &\bf\B             3.2 \\ 
\large              time pilot &      5,925.0 &\B     5,947.0 &         3,972.0 &\bf\B         8,540.0 \\ 
\large               tutankham &        167.7 &\bf\B    186.7 &            81.4 &                147.4 \\ 
\large               up n down &      9,082.0 &       8,456.0 &\B      19,533.0 &\bf\B        36,936.0 \\ 
\large                 venture &\bf   1,188.0 &         380.0 &           244.5 &                 80.0 \\ 
\large           video pinball &     17,298.0 &\B    42,684.0 &         9,783.9 &\bf\B       188,604.4 \\ 
\large           wizard of wor &      4,757.0 &       3,393.0 &         2,733.6 &\bf\B        40,780.0 \\ 
\large                  zaxxon &      9,173.0 &       4,977.0 &         8,204.4 &\bf\B        18,700.0 \\ 
\midrule
\large         \# $\geq$ Human &          n/a &   23 (46.9\%) &     18 (36.7\%) &\bf       25 (51.0\%) \\ 
\large    \# $\geq$ 75\% Human &          n/a &   27 (55.1\%) &     22 (44.8\%) &\bf       29 (59.1\%) \\ 
\midrule
\large         \# best in game &\bf 16 (32.6\%) & 12 (24.4\%) &      6 (12.2\%) &          15 (30.6\%) \\ 
\bottomrule
\end{tabular}
  }
  \caption{Scores for RAS Rollout IW(1) with a time budget of half a second vs.\ Human, DQN,
    and shallow RL algorithm using Blob-PROST features on 49 games.
    Best overall entries across rows are shown in bold, and scores as good as the human player
    are highlighted in red.
  }
  \label{table:narrow}
\end{table}

Table~\ref{table:big2} shows results for the human player, IW(1) using RAM, and
IW(1) using the screen, and three versions of Rollout IW(1) with time budgets
of 0.5 and 32 seconds. The table highlights in bold the entries that are best
across each row, and in red the those that are better than human playing (except
for IW(1) over RAM).
Overall, for a fixed time budget, RAS Rollout IW(1) is better than RA Rollout IW(1),
and this one is better than (plain) Rollout IW(1).
The risk-averse versions of Rollout IW(1) are the only IW(1) algorithms that
achieve a positive score for the challenging Montezuma's Revenge, in 0.5 seconds.
With 32 seconds, both RA and RAS Rollout IW(1) achieve more than 1,000 points,
a good score only worse than the human player that obtains 4,367 points.
Also, interestingly, IW(1) and the variants of Rollout IW(1) running over
the screen states often perform better than IW(1) running until completion over
the more informed RAM states.

\begin{figure}[t]
  \centering
  \includegraphics[width=\columnwidth]{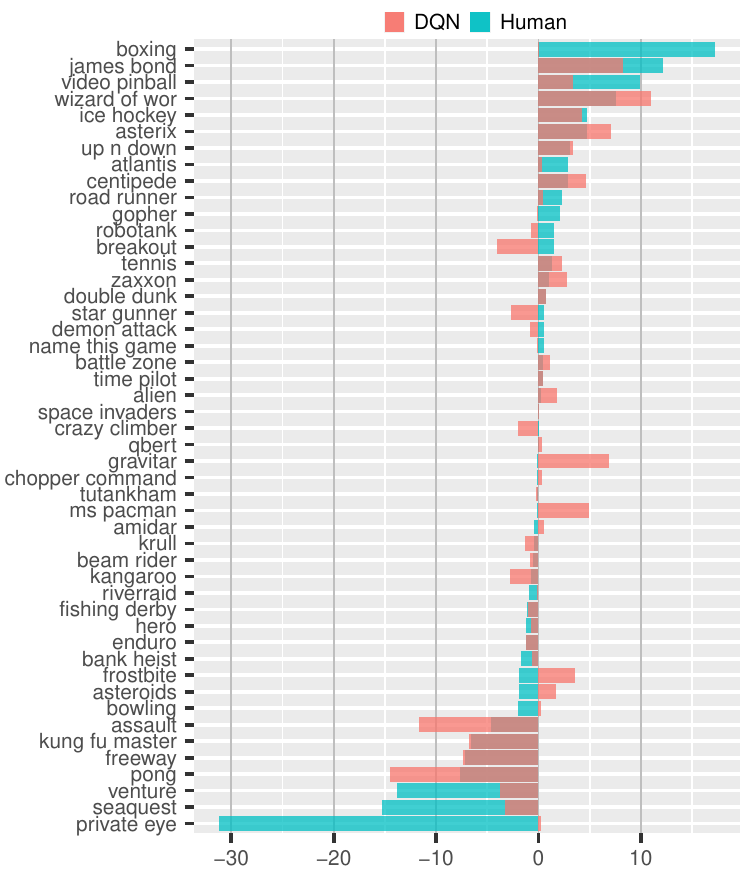}
  \caption{Comparison of RAS Rollout IW(1) with time budget of 0.5 seconds with
    the human player and DQN using relative scores over 48 games.
    Blue bars depict RAS Rollout IW(1) scores ($r)$ vs.\ human scores ($h$), while
    red bars, RAS Rollout IW(1) scores vs.\ DQN scores ($d$). The bar has positive length $(r-h)/h$
    when $r>h$, and negative length $(h-r)/r$ when $r<h$. Bars have zero length when $r=h$.
    Same for $d$ in place of $h$.
  }
  \label{fig:110:bars:half}
\end{figure}

Finally, Figure~\ref{fig:110:bars:half} shows a comparison of RAS Rollout IW(1)
with a time budget of half a second in relation to the human player and DQN.
Blue bars in the chart compare RAS Rollout IW(1) scores ($r$) and human scores
($h$), while red bars compare RAS Rollout IW(1) scores vs.\ DQN scores ($d$). 
The bar has positive length $(r-h)/h$ when $r>h$ (Rollout better), and negative
length $(h-r)/r$ when $r<h$ (Human better).  When $r=h=0$, the bar has zero length.
Similar relations hold for $d$ in place of $h$. 
Games that result in infinite sized bars are excluded. This only happens for
Montezuma's Revenge as the scores for DQN and RAS Rollout IW(1) are
0 and 100 respectively.

\begin{table*}[p]
  \centering
  \resizebox{\textwidth}{!}{
\newcommand{\B}{\color{red}}
\begin{tabular}{@{}rrrrr@{}rr@{}r@{}rr@{}r@{}rr@{}r@{}rr@{}}
\toprule
                             &&            &       IW(1) &&       \multicolumn{2}{c}{IW(1)} &&\multicolumn{2}{c}{Rollout IW(1)} &&\multicolumn{2}{c}{RA Rollout IW(1)} &&\multicolumn{2}{c}{RAS Rollout IW(1)} \\
\cline{6-7}
\cline{9-10}
\cline{12-13}
\cline{15-16}
                        Game &&      Human &         RAM &\textcolor{white}{x}&  budget .5s &  budget 32s &\textcolor{white}{x}&  budget .5s &   budget 32s &\textcolor{white}{x}&  budget .5s &   budget 32s &\textcolor{white}{x}&  budget .5s &   budget 32s \\
\midrule
                       alien &&    6,875.0 &\bf 25,634.0 &&        1,316.0 &\B     14,010.0 &&        4,238.0 &\B       6,896.0 &&\B      7,170.0 &\B      13,454.0 &&\B      8,550.0 &\B      19,354.0 \\
                      amidar &&    1,676.0 &     1,377.0 &&           48.0 &        1,043.2 &&          659.8 &\B       1,698.6 &&        1,049.2 &\bf\B    1,794.0 &&        1,161.0 &         1,609.0 \\
                     assault &&\bf 1,496.0 &       953.0 &&          268.8 &          336.0 &&          285.6 &           319.2 &&          336.0 &           327.6 &&          264.6 &           281.4 \\
                     asterix &&    8,503.0 &   153,400.0 &&        1,350.0 &\bf\B 262,500.0 &&\B     45,780.0 &\B      66,100.0 &&\B     46,100.0 &\B      67,100.0 &&\B     48,700.0 &\B      87,600.0 \\
                   asteroids &&   13,157.0 &\bf 51,338.0 &&          840.0 &        7,630.0 &&        4,344.0 &         7,258.0 &&        4,698.0 &         6,836.0 &&        4,486.0 &         7,344.0 \\
                    atlantis &&   29,028.0 &\bf159,420.0 &&\B     33,160.0 &\B     82,060.0 &&\B     64,200.0 &\B     151,120.0 &&\B    122,220.0 &\B     134,660.0 &&\B    113,460.0 &\B     134,660.0 \\
                  bank heist &&      734.4 &       717.0 &&           24.0 &\B        739.0 &&          272.0 &\B         865.0 &&          242.0 &\B       1,323.4 &&          268.0 &\bf\B    2,179.0 \\
                 battle zone &&   37,800.0 &    11,600.0 &&        6,800.0 &       14,800.0 &&\B     39,600.0 &\B     414,000.0 &&\B     74,600.0 &\B     455,800.0 &&\B     56,200.0 &\bf\B  509,400.0 \\
                  beam rider &&    5,775.0 &\bf  9,108.0 &&          715.2 &        1,530.4 &&        2,188.0 &         2,464.8 &&        2,552.8 &         5,367.2 &&        3,729.2 &         4,921.2 \\
                     berzerk &&        n/a &\bf  2,096.0 &&          280.0 &        1,318.0 &&          644.0 &           862.0 &&        1,208.0 &         1,454.0 &&          966.0 &         1,640.0 \\
                     bowling &&\bf   154.8 &        69.0 &&           30.6 &           49.2 &&           47.6 &            45.8 &&           44.2 &            49.0 &&           51.6 &            48.0 \\
                      boxing &&        4.3 &\bf    100.0 &&\B         99.4 &\B         79.0 &&\B         75.4 &\B          79.4 &&\B         99.2 &\bf\B      100.0 &&\B         78.6 &\B          80.2 \\
                    breakout &&       31.8 &\bf    384.0 &&            1.6 &\B         56.0 &&\B         82.4 &\B          36.0 &&\B         86.2 &\B         336.4 &&\B         79.8 &\B         370.0 \\
                   centipede &&   11,963.0 &    99,207.0 &&\B     88,890.0 &\bf\B 143,275.4 &&\B     36,980.2 &\B      65,162.6 &&\B     56,328.0 &\B      92,353.0 &&\B     46,661.4 &\B      84,226.0 \\
             chopper command &&    9,882.0 &    10,980.0 &&        1,760.0 &        1,800.0 &&        2,920.0 &         5,800.0 &&        9,820.0 &\B      11,240.0 &&        8,900.0 &\bf\B   33,220.0 \\
               crazy climber &&   35,411.0 &    36,160.0 &&       16,780.0 &\bf\B  44,340.0 &&\B     39,220.0 &\B      43,960.0 &&\B     40,440.0 &\B      38,460.0 &&\B     38,120.0 &\B      42,720.0 \\
                    defender &&        n/a &         n/a &&      362,010.0 &\bf   430,010.0 &&      373,010.0 &       256,010.0 &&      371,010.0 &       374,010.0 &&      298,010.0 &       387,010.0 \\
                demon attack &&    3,401.0 &    20,116.0 &&          106.0 &\bf\B  23,619.0 &&        2,780.0 &\B       9,996.0 &&\B      6,958.0 &\B      10,753.0 &&\B      5,201.0 &\B       9,898.0 \\
                 double dunk &&      -15.5 &       -14.0 &&          -22.0 &          -22.4 &&\B          3.6 &\bf\B       20.0 &&\B          3.2 &\B          19.6 &&\B         -4.0 &\B          16.0 \\
             elevator action &&        n/a &\bf 13,480.0 &&        1,080.0 &            0.0 &&            0.0 &             0.0 &&            0.0 &             0.0 &&            0.0 &         1,300.0 \\
                      enduro &&      309.6 &\bf    500.0 &&            2.6 &          229.2 &&          169.4 &\B         359.4 &&          145.8 &\B         381.0 &&          137.4 &\B         330.8 \\
               fishing derby &&        5.5 &\bf     30.0 &&          -83.8 &          -39.0 &&          -68.0 &           -16.2 &&          -77.0 &           -50.2 &&          -61.8 &           -53.0 \\
                     freeway &&       29.6 &\bf     31.0 &&            0.6 &           25.0 &&            2.8 &            12.6 &&            2.0 &            11.2 &&            3.6 &            10.0 \\
                   frostbite &&    4,335.0 &       902.0 &&          106.0 &          182.0 &&          220.0 &\B       5,484.0 &&          146.0 &\bf\B    6,398.0 &&        1,494.0 &\B       5,970.0 \\
                      gopher &&    2,321.0 &    18,256.0 &&        1,036.0 &\bf\B  18,472.0 &&\B      7,216.0 &\B      13,176.0 &&\B      8,388.0 &\B      13,144.0 &&\B      7,256.0 &\B      11,840.0 \\
                    gravitar &&    2,672.0 &     3,920.0 &&          380.0 &        1,630.0 &&        1,630.0 &\B       3,700.0 &&        1,660.0 &\B       5,130.0 &&        2,410.0 &\bf\B    5,540.0 \\
                        hero &&   25,673.0 &    12,985.0 &&        2,034.0 &        7,432.0 &&       13,709.0 &\B      28,260.0 &&       11,377.0 &        24,072.0 &&       11,480.0 &\bf\B   29,708.0 \\
                  ice hockey &&        0.9 &\bf     55.0 &&          -13.6 &           -7.0 &&           -6.0 &             6.6 &&          -12.4 &            -2.6 &&\B          5.2 &\B          18.2 \\
                  james bond &&      406.7 &\bf 23,070.0 &&           40.0 &          180.0 &&\B        450.0 &\B      22,250.0 &&\B     10,760.0 &\B      12,656.0 &&\B      5,340.0 &\B      12,345.0 \\
                      kaboom &&        n/a &         n/a &&          127.0 &          178.2 &&           87.4 &           159.6 &&           23.2 &           151.6 &&           39.8 &\bf        255.6 \\
                    kangaroo &&    3,035.0 &\bf  8,760.0 &&          160.0 &\B      3,820.0 &&        1,080.0 &\B       5,780.0 &&        1,880.0 &\B       4,600.0 &&        1,800.0 &\B       5,280.0 \\
                       krull &&    2,395.0 &\bf  6,030.0 &&\B      3,206.8 &\B      5,611.8 &&        1,892.8 &         1,151.2 &&        2,091.8 &         2,219.8 &&        1,645.2 &\B       2,837.0 \\
              kung fu master &&   22,736.0 &\bf 63,780.0 &&          440.0 &        8,980.0 &&        2,080.0 &        14,920.0 &&        2,620.0 &\B      26,540.0 &&        2,980.0 &\B      24,300.0 \\
         montezuma's revenge &&\bf 4,367.0 &         0.0 &&            0.0 &            0.0 &&            0.0 &             0.0 &&          100.0 &         1,620.0 &&          100.0 &         1,080.0 \\
                   ms pacman &&   15,693.0 &    21,695.0 &&        2,578.0 &\B     20,622.8 &&        9,178.4 &\B      19,667.0 &&       15,115.0 &\bf\B   23,033.0 &&       13,746.8 &\B      21,833.0 \\
              name this game &&    4,076.0 &     9,354.0 &&\B      7,070.0 &\bf\B  13,478.0 &&\B      6,226.0 &\B       5,980.0 &&\B      6,558.0 &\B       6,870.0 &&\B      6,128.0 &\B       6,820.0 \\
                     phoenix &&        n/a &         n/a &&        1,266.0 &        5,550.0 &&        5,750.0 &\bf      7,636.0 &&        6,790.0 &         7,460.0 &&        5,386.0 &         7,570.0 \\
                     pitfall &&        n/a &         n/a &&\bf        -8.6 &          -92.2 &&          -81.4 &          -130.8 &&         -302.8 &          -723.4 &&         -814.6 &          -692.4 \\
                        pong &&        9.3 &\bf     21.0 &&          -20.8 &            0.8 &&           -7.4 &\B          17.6 &&           -4.2 &\B          15.2 &&           -1.4 &\B          18.2 \\
                 private eye &&\bf69,571.0 &       -99.0 &&        2,690.8 &         -526.4 &&         -322.0 &         3,157.2 &&         -480.0 &          -300.0 &&        2,160.0 &          -340.0 \\
                       qbert &&   13,455.0 &     3,705.0 &&          515.0 &\B     16,505.0 &&        3,375.0 &         8,390.0 &&\B     15,970.0 &\B      26,875.0 &&\B     14,160.0 &\bf\B   40,350.0 \\
                   riverraid &&   13,513.0 &     5,694.0 &&          664.0 &        7,042.0 &&        6,088.0 &         8,156.0 &&        6,288.0 &\B      14,700.0 &&        7,138.0 &\bf\B   15,302.0 \\
                 road runner &&    7,845.0 &\bf 94,940.0 &&          200.0 &            0.0 &&        2,360.0 &\B      37,080.0 &&\B     31,140.0 &\B      44,020.0 &&\B     25,780.0 &\B      62,960.0 \\
                    robotank &&       11.9 &\bf     68.0 &&            3.2 &\B         32.8 &&\B         31.0 &\B          52.6 &&\B         31.2 &\B          47.4 &&\B         30.0 &\B          51.8 \\
                    seaquest &&\bf20,182.0 &    14,272.0 &&          168.0 &          356.0 &&          980.0 &        10,932.0 &&        2,312.0 &        13,472.0 &&        1,236.0 &         9,846.0 \\
                      skiing &&        n/a &         n/a &&      -16,511.0 &      -15,962.0 &&      -15,738.8 &       -16,477.0 &&      -16,006.8 &\bf    -15,244.2 &&      -15,806.6 &       -15,473.6 \\
                     solaris &&        n/a &         n/a &&        1,356.0 &\bf     2,300.0 &&          700.0 &         1,040.0 &&        1,704.0 &           692.0 &&        1,620.0 &         1,728.0 \\
              space invaders &&    1,652.0 &     2,877.0 &&          280.0 &\B      1,963.0 &&\B      2,628.0 &\B       1,980.0 &&        1,149.0 &\B       2,533.0 &&\B      1,812.0 &\bf\B    4,362.0 \\
                 star gunner &&   10,250.0 &     1,540.0 &&          840.0 &        1,340.0 &&\B     13,360.0 &\B      15,640.0 &&\B     14,900.0 &\B      16,460.0 &&\B     15,960.0 &\bf\B   17,160.0 \\
                      tennis &&       -8.9 &\bf     24.0 &&          -23.4 &          -22.2 &&          -18.6 &\B          -2.2 &&\B         -5.4 &\B          14.8 &&\B          3.2 &\B          -3.4 \\
                  time pilot &&    5,925.0 &\bf 35,000.0 &&        2,360.0 &        5,740.0 &&\B      7,640.0 &\B       8,140.0 &&        3,540.0 &         5,480.0 &&\B      8,540.0 &         5,480.0 \\
                   tutankham &&      167.7 &       172.0 &&           71.2 &\B        172.4 &&          128.4 &\B         184.0 &&          135.6 &\bf\B      191.6 &&          147.4 &\B         191.2 \\
                   up n down &&    9,082.0 &\bf110,036.0 &&          928.0 &\B     62,378.0 &&\B     36,236.0 &\B      44,306.0 &&\B     34,668.0 &\B      39,964.0 &&\B     36,936.0 &\B      41,056.0 \\
                     venture &&    1,188.0 &\bf  1,200.0 &&            0.0 &          240.0 &&            0.0 &            80.0 &&           60.0 &           500.0 &&           80.0 &           120.0 \\
               video pinball &&   17,298.0 &   388,712.0 &&\B     28,706.4 &\bf\B 441,094.2 &&\B    203,765.4 &\B     382,294.8 &&\B    216,468.6 &\B     378,815.4 &&\B    188,604.4 &\B     375,073.0 \\
               wizard of wor &&    4,757.0 &\bf121,060.0 &&\B      5,660.0 &\B    115,980.0 &&\B     37,220.0 &\B      73,820.0 &&\B     43,860.0 &\B      84,660.0 &&\B     40,780.0 &\B      75,380.0 \\
                yars revenge &&        n/a &         n/a &&        6,352.6 &       10,808.2 &&        5,225.4 &         9,866.4 &&        7,848.8 &\bf     23,261.8 &&        3,647.8 &        10,523.6 \\
                      zaxxon &&    9,173.0 &    29,240.0 &&            0.0 &\B     15,080.0 &&\B      9,280.0 &\B      22,880.0 &&\B     15,500.0 &\B      34,180.0 &&\B     18,700.0 &\bf\B   38,700.0 \\
\midrule
             \# $\geq$ Human &&        n/a &         n/a &&     7 (14.2\%) &    22 (44.8\%) &&    19 (38.7\%) &     34 (69.3\%) &&    22 (44.8\%) &     35 (71.4\%) &&    25 (51.0\%) &\bf  37 (75.5\%) \\
        \# $\geq$ 75\% Human &&        n/a &         n/a &&     7 (14.2\%) &    24 (48.9\%) &&    22 (44.8\%) &     34 (69.3\%) &&    26 (53.0\%) &     39 (79.5\%) &&    29 (59.1\%) &\bf  40 (81.6\%) \\
\midrule
             \# best in game &&  5 (8.6\%) &\bf  24 (41.3\%) &&  1 (1.7\%) &     9 (15.5\%) &&      0 (0.0\%) &       2 (3.4\%) &&      0 (0.0\%) &      7 (12.0\%) &&      0 (0.0\%) &     11 (18.9\%) \\
\bottomrule
\end{tabular}
  }
  \caption{
    Scores over 58 games for Human, IW(1) over RAM, and IW(1) and three versions of Rollout IW(1) with time budgets
    of 0.5 and 32 seconds. Entries marked as n/a are not available, those highlighted in bold are best scores across
    rows, and those highlighted in red are scores as good as human playing (except for IW(1) over RAM).
    The 9 games for which there are no reports for the human player are not considered when comparing the 
    different algorithms with the human player.
  }
  \label{table:big2}
\end{table*}

\section{Discussion}

We have introduced a new width-based search algorithm, Rollout IW($k$), whose
theoretical properties are similar to the ones for IW($k$), and have shown that
it plays Atari games from the screen pixels, almost in real-time at the level
of humans and some of the best reinforcement learning algorithms.
These are the first results for planning in the Atari games from the screen, and
they make use of the B-PROST set of visual features defined in \cite{shallow}.
Interestingly, the results compare favorably with those obtained by planning over
the memory states in many games as well. More recent RL algorithms for the ALE
environment are reviewed and combined in \cite{silver:rainbow}, while recent work
concerned with planning with images is reported in \cite{fukunaga:planning-images}.



Two key differences between the proposed planning approach and reinforcement learning
methods is that the latter can deal naturally with stochastic actions and result in
closed-loop controllers that do not require any lookahead.
One challenge for the future is to extend width-based methods such as IW(1) and Rollout IW(1)
to stochastic settings, such as the ``noisy'' version of the Atari games supported in the
latest ALE release \cite{machado:ale}. A second challenge is to use planners to
obtain closed-loop policies that do not require any lookahead. One approach to do this 
is through imitation learning with the planner playing the role of the teacher. 
A more sophisticated iterated approach, akin to a pointwise version of approximate policy iteration, 
has been recently used in the Alpha Go Zero program that learns to play Go at superhuman level,
from self-play alone, using a MCTS planner \cite{alpha-zero-go}.
Another theme for future work is the study of alternative sets of general visual
features and how they relate to the theoretical width of games and the performance
of the algorithms.

\medskip
\noindent\textbf{Acknowledgements:}
We thank the ALE team at the University of Alberta for the challenge and the software environment,
Guillem Franc\`es for helping Wilmer get started in this project, and
Nir Lipovetzky for useful comments.
H. Geffner is partially funded by grant TIN-2015-67959-P, MINECO, Spain.

\bibliographystyle{aaai}
\bibliography{control}

\end{document}